\documentclass[10pt,twocolumn,letterpaper]{article}
\usepackage[accsupp]{axessibility} 
\usepackage{iccv}
\usepackage{times}
\usepackage{epsfig}
\usepackage{graphicx}
\usepackage{amsmath}
\usepackage{amssymb}
\usepackage{multirow}
\usepackage{bm}
\usepackage{colortbl}
\usepackage[dvipsnames]{xcolor}
\usepackage{xcolor,colortbl}
\usepackage{amsmath}
\usepackage{enumitem}

\iccvfinalcopy

\definecolor{Gray}{gray}{0.85}
\definecolor{LightCyan}{rgb}{0.88,1,1}

\usepackage[breaklinks=true,bookmarks=false]{hyperref}
\newcommand{\name}{\textbf{RepFusion}}

\def\iccvPaperID{10685} 

\ificcvfinal\fi

\begin{document}

\title{Diffusion Model as Representation Learner}

\author{Xingyi Yang \quad \quad Xinchao Wang\thanks{Corresponding Author.} \\
National University of 
Singapore\\
{\tt\small xyang@u.nus.edu, xinchao@nus.edu.sg}
}

\maketitle
\ificcvfinal\thispagestyle{empty}\fi

\begin{abstract}
Diffusion Probabilistic Models (DPMs) have recently demonstrated
impressive results on various generative tasks.
Despite its promises,
the learned representations of pre-trained DPMs, however,
have not been fully understood. 
In this paper, we conduct an in-depth 
investigation of the representation power of DPMs,
and propose a novel knowledge transfer method that  
leverages the knowledge acquired by generative DPMs 
for recognition tasks. 
Our study begins by examining the feature space of DPMs, 
revealing that DPMs are inherently denoising autoencoders
that balance the   representation learning
with regularizing model capacity. To this end, we introduce a novel knowledge transfer paradigm named \textbf{RepFusion}. Our paradigm  extracts representations at different time steps from off-the-shelf DPMs and dynamically employs them as supervision for student networks, in which the optimal time is determined through reinforcement learning. We evaluate our approach on several image classification, semantic segmentation, and landmark detection benchmarks, and demonstrate that it outperforms state-of-the-art methods. Our results uncover the potential of DPMs as a powerful tool for representation learning and provide insights into the usefulness of generative models beyond sample generation. The code is available at \url{https://github.com/Adamdad/Repfusion}.
\end{abstract}


\section{Introduction}
In the ever-evolving landscape of machine learning, generative models have emerged as a captivating approach to tackle the intricacies of data distributions. Among these marvels, Diffusion Probabilistic Models (DPMs) stand tall, boasting a remarkable prowess in producing realistic and diverse samples. Powered by the elegent design of diffusion, these models elegantly transform a humble noise into target data distribution, unfurling a breathtaking array of variations and unyielding fidelity in their artistic creations.

Although the generative ability of DPMs has been extensively studied, their potential for representation learning has not been fully explored. Recent research has demonstrated that the diffusion models already have a semantic latent space\cite{kwon2022diffusion}, and could be extended to tasks like controllable image generation~\cite{preechakul2021diffusion}, representation learning~\cite{abstreiter2021diffusion,mittal2022points} and image segmentation~\cite{baranchuk2021label,xu2023odise}, albeit through complicated model modifications. Nevertheless, the usefulness of the learned feature space of DPMs remain unclear. In this work, we intend to close the gap between generative DPM and its capability in representation learning by answering the following question: \textit{Can the representation learned by DPMs be effectively reused for recognition tasks?} 

\begin{figure}
    \centering
    \includegraphics[width=\linewidth]{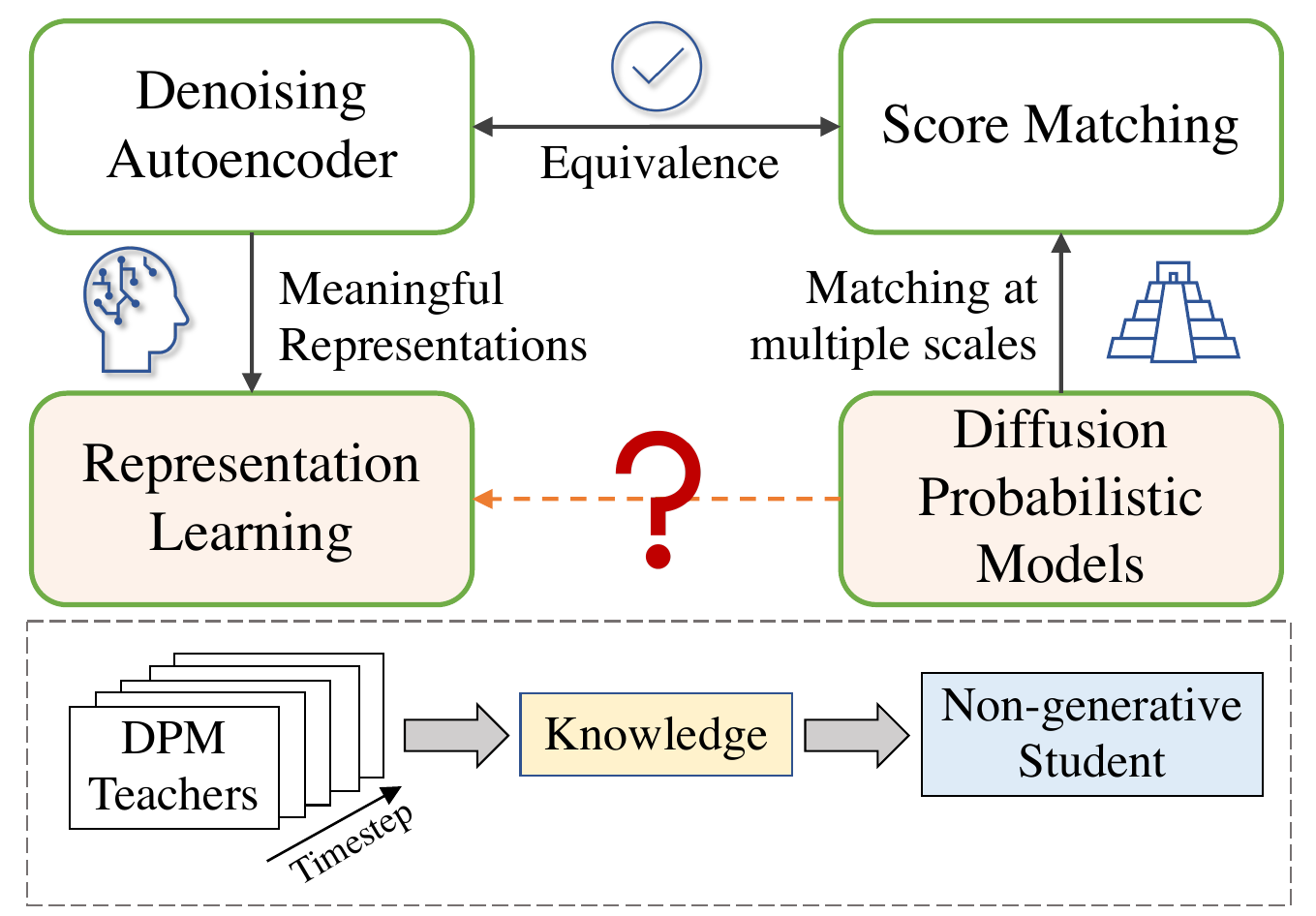}
    \caption{Illustration of our core idea on establishing a link between representation learning and diffusion models. Building upon this connection, we propose to utiliz the knowledge acquired from pre-trained DPMs. This knowledge is selectively distilled into a student network, empowering it to carry out recognition tasks with enhanced proficiency. }
    \label{fig:concepts}
    \vspace{-2mm}
\end{figure}

We answer this question by first analyzing the  inherent relationship between the diffusion model and the standard auto-encoder, as shown in Figure~\ref{fig:concepts}.
 Conceptually, DPMs are designed to predict noise from perturbed input distributions. Essentially, DPMs can be viewed as \emph{denoising auto-encoders}~(DAEs)~\cite{vincent2010stacked,vincent2008extracting} with varying denoising scales. DAEs have been well-established as a powerful technique for self-supervised learning, which captures the underlying structures of data. Its latent features are useful for downstream tasks. On another note, DPMs are commonly known as score-matching models~\cite{song2019generative}, which model the gradient of the log probability density with a sequence of intermediate latent variables. By connecting score-matching with DAEs~\cite{vincent2011connection}, DPMs should also be considered an extension of DAE. It is, therefore, intuitive that DPM produces meaningful encoding from its input. 

Although the representational power of the diffusion model has well-established theoretical foundations, leveraging the off-the-self DPM for non-generative tasks poses significant challenges. Primarily, these models are parameterized as time-conditioned Unet~\cite{ronneberger2015u,dhariwal2021diffusion,song2021scorebased}, a specialized structure unsuited for tasks such as classification and object detection. Secondly, existing DPMs are computationally heavy~\cite{yang2022diffusion}, making it difficult to use the original model for discriminative tasks without substantial modifications. Another obstacle arises from the fact that a single DPM can be perceived as a composition of networks, each indexed by its input timestep. Determining the suitable time-step remains non-trivial. Consequently, the representations learned by DPMs do not readily benefit other non-generative tasks.
 
In this paper, our primary objective is to reuse the knowledge encoded in the DPM for recognition tasks~\cite{deng2009imagenet}. {To achieve this, we look into the mathematical formulation of DPM's latent space, and show that DPMs can be seen as DAEs that strike a balance between learning meaningful features and regularizing the model capacity.} Building on this insight, we propose a novel knowledge transfer approach, termed~\name. Our approach utilizes the knowledge distillation techniques to transfer the representation learning capability of the trained generative models to  improve discriminative tasks. Specifically, we dynamically extract intermediate representations at different time steps and use them as auxiliary supervision  for student networks. To determine the optimal time selection, we measure the informativeness of a given representation and optimize it through the REINFORCED algorithm~\cite{williams1992simple}. {Moreover, the reinforced time-steps selection is aligned with the derived property of DPMs, thereby providing a mechanism for adapting to different downstream tasks and increasing the flexibility and generalizability of our approach.} 

Our experiments demonstrate that~\name~consistently improves performance on several image classification, semantic segmentation, and landmark detection benchmarks, indicating the powerful representation learning capability of DPMs. These findings shed light on the potential utility of generative models beyond their traditional use in sample generation and highlight the opportunities for exploring pre-trained models in representation learning.

To sum up, our contributions can be divided into three main parts:
\begin{enumerate}
\item We investigate the potential of repurposing diffusion models for representation learning, an area that has been relatively unexplored in prior research.
\item By establishing the relationship between DPMs and denoising auto-encoders, we verify the statistical and empirical properties of features extracted from DPMs.
\item We introduce a novel knowledge distillation approach called~\name, which utilizes pre-trained DPMs to enhance recognition tasks. Extensive evaluations on image classification, segmentation, and landmark detection benchmarks demonstrate the effectiveness of DPMs as powerful tools for representation learning.
\end{enumerate}
\section{Related Work}

\noindent\textbf{Denoising Autoencoders.} 
Autoencoders are a type of neural network that can learn to reconstruct their input at the output. 
Denoising autoencoder (DAE) is a variant of autoencoder that recover clean output from corrupted input. With the denoising objective, DAE has been one of the dominant approaches for unsupervised representations learning~\cite{vincent2010stacked,vincent2008extracting,he2022masked,lewis2019bart}. Several studies have demonstrated that adjusting the noise scale~\cite{geras2014scheduled,chandra2014adaptive,zhang2018convolutional} improves the features. Apart from learning representation, DAE has been recognized as a generative model by matching the score function~\cite{vincent2011connection,song2019generative, NIPS2013_559cb990,kingma2010regularized}. This paper reframes modern diffusion models as auto-encoders to make their latent features more suitable for recognition tasks.\\
\noindent\textbf{Diffusion Probabilistic Models.} DPMs~\cite{ho2020denoising} have emerged as a cutting-edge approach for generating high-quality samples, with applications like conditioned image generation~\cite{zhang2023adding,ramesh2022hierarchical,DBLP:conf/nips/SahariaCSLWDGLA22}, 3D generation~\cite{poole2023dreamfusion}, video synthesis~\cite{ho2022video} and domain generalization~\cite{Yu_2023_CVPR}. DPMs are essentially score-based models~\cite{vincent2011connection,song2021scorebased} match the score functions over multiple scales~\cite{song2019generative}. The objective of this study is to investigate the potential of diffusion models in learning representation, which has not received much attention before.\\
\noindent\textbf{Learn Representation from Generative Models} Generative models, such as diffusion-based models have been found to learn meaningful semantics~\cite{preechakul2021diffusion,kwon2022diffusion}, with applications on semantic segmentation~\cite{xu2023odise}, correspondence~\cite{luo2023dhf,Tang2023EmergentCF} and image editing~\cite{shi2023dragdiffusion}. Others endeavors to learn from model generated data~\cite{DBLP:journals/corr/abs-1711-04340,DBLP:journals/corr/abs-2305-12954,DBLP:conf/icra/XiangXXZZM23,DBLP:conf/iclr/HeS0XZTBQ23}. However, current approaches~\cite{baranchuk2021label} require complex modifications of the model. In contrast, our paper takes a novel approach by distilling well-trained DPMs for general representation learning.\\
\noindent\textbf{Model Reuse and Knowledge Distillation.} 
The popularity of pre-trained models creates a growing demand for reusing model~\cite{pmlr-v202-wang23aa,jing2021amalgamating,jing2022learning,yang2022factorizing,yang2022deep,SGFormerICCV23} and data~\cite{DBLP:conf/nips/Liu0YYW22,Liu_2023_CVPR} to enhance performance and minimize costs~\cite{wang2023learning,fang2023depgraph,fang2023structural,Xinyin2023structural}.
Knowledge distillation (KD)~\cite{Hinton2015DistillingTK,yang2020CVPR,tian2019contrastive,li2023diverse} emerges as a promising solution, facilitating knowledge transfer from teacher models to students, benefiting DPMs with reduced sampling time~\cite{Meng_2023_CVPR,salimans2022progressive,Meng_2023_CVPR} and improved model efficiency~\cite{yang2022diffusion}. Our research pioneers the exploration of knowledge distillation from diffusion-based models for recognition tasks.
\section{Diffusion Models are Auto-Encoders}
\label{sec:dpm_are_ae}
In this section, we review DPM's formulation and demonstrate its connection to DAEs in terms of representation learning. Our main insight is that DPM behaves similarly to a regularized auto-encoder, with its representation characteristics being controlled by noise scales.
\subsection{Recap on Diffusion Probabilistic Models}
The diffusion model is a probabilistic generative model designed for denoising by systematically reversing a progressive noising process. It initiates with clean data denoted as $\mathbf{x}_0\sim q(\mathbf{x}_0)$ and iteratively applies Gaussian noise with zero mean and variance $\beta_t$ for a total of $T$ steps, progressively generating a noisy version at each step,
\begin{align}
    q(\mathbf{x}_t|\mathbf{x}_{t-1})=\mathcal{N}(\mathbf{x}_{t};\sqrt{1-\beta_{t}}\mathbf{x}_{t-1}, \beta_{t}\mathbf{I})
\end{align}
where $t\in [1, T]$ and $0<\beta_{1:T}<1$ denote the noise scale scheduling. This process converges to a Gaussian white noise distribution, i.e., $\mathbf{x}_T \to \mathcal{N}(0, \mathbf{I})$. 
Sampling from the noise-perturbed distribution $q(\mathbf{x}_t)=\int q(\mathbf{x}_{1:t}|\mathbf{x}_0) d \mathbf{x}_{1:t-1}$ requires numerical integration over the steps, while the use of Gaussian parametrization enables the generation of an arbitrary time-step $\mathbf{x}_t$ through a closed-form formulation
\begin{align}
    \mathbf{x}_t = \sqrt{\bar{\alpha}_t} \mathbf{x}_0 + \sqrt{1-\bar{\alpha}_t} \bm{\epsilon}, \quad \text{where} \quad \bm\epsilon \sim \mathcal{N}(0, \mathbf{I})
\label{eq:xt}
\end{align}
where $\alpha_t = 1-\beta_t$ and $\bar{\alpha}_t = \prod_{s=1}^t \alpha_s$. The reverse process is parameterized as a time-conditioned denoising neural network $\mathbf{s}(\mathbf{x}, t;\bm \theta)$ with $p_{\bm \theta}(\mathbf{x}_{t-1}|\mathbf{x}_t)=\mathcal{N}(\mathbf{x}_{t-1}; \frac{1}{\sqrt{1-\beta_t}}(\mathbf{x}_t+\beta_t \mathbf{s}(\mathbf{x}_t, t;\bm \theta)), \beta_t \mathbf{I})$. This neural network is trained to minimize re-weighted evidence lower bound (ELBO) that fits the noise
\begin{align}
    \mathcal{L}_{\text{DDPM}} &= \mathbb{E}_{t,\mathbf{x}_0, \bm\epsilon} \Big[||\bm\epsilon - \bm s(\mathbf{x}_t,t;\bm \theta)||_2^2\Big]\label{eq:loss_ddpm}
\end{align}
With a trained denoiser $\mathbf{s}(\mathbf{x}, t;\bm \theta^*)$, we are able to generate the data by solving a reverse process. 

\noindent\textbf{Connecting DAE with Diffusion Model.} The relation between diffusion model and DAE lies at the core of our paper. In essence, diffusion models are created through a gradual denoising procedure, akin to a cascade of DAEs, with the only distinction on the input $t$ and shared parameters.

A compelling demonstration of this connection can be traced back to the advent of \textit{score-based} generative model~\cite{hyvarinen2005estimation,vincent2011connection,bengio2012implicit,alain2013regularized,NIPS2013_559cb990,alain2014regularized}. In that sense, the denoising auto-encoders involves fitting the derivative of the log-likelihood of the data, which is known as~\textit{score matching}. The connection is further reinforced by a recent study~\cite{song2021scorebased} that diffusion models learn to generate data by matching scores. This profound insight provides a deeper understanding of the intrinsic relationship between these two powerful models, paving the way for our later discussion.

\subsection{Probing the Representations in DPMs}
\label{sec:prob_representation}



Despite the connection between DPM and DAE, we typically ignore internal latent representation of DPM. This section examines the features learned by DPM, particularly the characteristics of the latent space at various time steps.

An autoencoder (AE) consists of an encoder $E:\mathbb{R}^{L} \to \mathbb{R}^{d}$ and a decoder $D:\mathbb{R}^{d} \to \mathbb{R}^{L}$, where $d<L$. The encoder maps the noisy input $\mathbf{x}'=\mathbf{x}+\sigma^2\bm \epsilon$ to a hidden space $\mathbf{h}$, and the decoder maps it back $\mathbf{x}'$ to make the prediction $\mathbf{y}$. We assume a linear AE with skip connection, which mimics the Unet~\cite{ronneberger2015u} used in DPMs. 
This encoder-decoder interaction is mathematically expressed as:
\begin{align}
    \mathbf{h}=W_E\mathbf{x}';\quad
    \mathbf{y}=W_D\mathbf{h}+W_S \mathbf{x}'
\end{align}
Here, $W_E\in \mathbb{R}^{d\times L}$, $W_D\in \mathbb{R}^{L\times d}$ and $W_S\in \mathbb{R}^{L\times L}$ are the encoding, decoding, and skip-connection matrices, respectively. This simplification enables us to examine the latent space properties of DPM in Proposition 1.

\noindent\textbf{Proposition 1.}\footnote{Due to the page limit, full derivation is in the supplementary material.} \textit{Assume a linear DPM with skip connection. For $\mathbf{x}_t = \sqrt{\bar{\alpha}} \mathbf{x}_0 + \sqrt{1-\bar{\alpha}} \bm{\epsilon}$ and $\mathbb{E}[\mathbf{x}_0] =0$, minimizing Eq.\ref{eq:loss_ddpm} is equivalent to minimize} 
{\small
\begin{align}
    \mathcal{L}(W_E,W_D, &W_S) = \mathbb{E}_{\mathbf{x}_0, \bm\epsilon }[ ||\bm\epsilon - (W_DW_E\mathbf{x}_t + W_S\mathbf{x}_t)||^2_2]\\
    &= \underbrace{\bar{\alpha}_t P^\top \Sigma_{xx} P }_{\text{Representation Learning}}+\underbrace{||I - \sqrt{1-\bar{\alpha}_t}P||^2_2}_{\text{Regularization}}\label{eq:1}
\end{align}
}
Here, $P =W_DW_E + W_S$, and $\Sigma_{xx}$ is the covariance matrix of $\mathbf{x}_0$. The expression consists of two terms: the first term is related to representation learning, while the second term is related to regularization.

\begin{enumerate}[align=left,labelwidth=\parindent,label=(\roman*)]
\itemsep0em 
    \item The first term ensures a meaningful latent space $\mathbf{h}$. For example, when $W_S=-I$, we get back to linear AE that encodes the data into principal components space~\cite{izenman1975reduced,bourlard1988auto}. \item The second term enforces spectral regularization~\cite{miyato2018spectral}, driving the diagonal values of the weight matrix $W_DW_E + W_S$ towards $\frac{1}{\sqrt{1-\bar{\alpha}_t}}$. This regularization encourages a compact latent space as more noise is added, aligning with the manifold hypothesis~\cite{cayton2005algorithms}.

\end{enumerate}



\noindent\textbf{Trade-off in DPMs.} Eq.~\ref{eq:1} indicates a trade-off between learning a meaningful representation and the penalize the spectrum of the optimal parameters. As $t$ gets larger and $\bar{\alpha}_t$ decreases, the model prioritizes the regularization term over the representation learning term. Consequently, the singular values of the learned representations tend to become smaller. if the model becomes overly regularized, its ability to learn informative features may become limited. Ideally, an optimal representation should exhibit uniform singular values close to 1, with a larger effective rank number ~\cite{roy2007effective} that prevents dimensional collapse~\cite{shi2023towards}. 

\noindent\textbf{Singular Values and Rank Number.}
To validate the trade-off in DPMs, we analyzed the singular values and effective rank number~\cite{roy2007effective} of embeddings extracted from well-trained DPMs\footnote{We feed clean images and timestep $t$ into the encoder and compute the statistics of the embedding vectors at the \textsc{mid-block} layer.} on the CIFAR10~\cite{Krizhevsky09learningmultiple} and MNIST~\cite{lecun2010mnist}. 

\begin{figure}
    \centering
    \includegraphics[width=\linewidth]{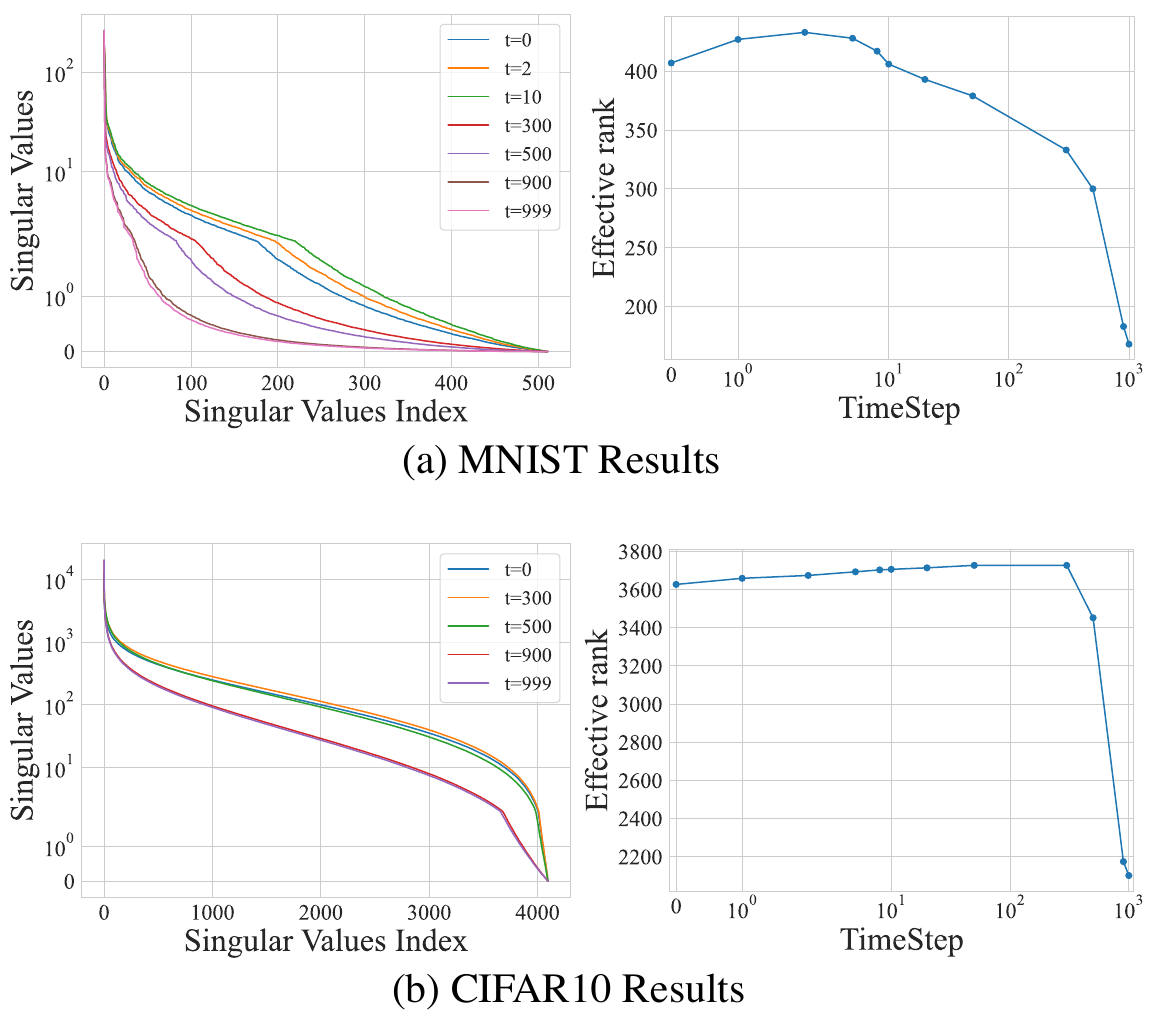}
    \caption{Comparison of (Left) the singular values and (Right) effective rank of learned features at different timestep on MNIST and CIFAR-10 datasets.}
    \label{fig:singular}
\end{figure}

As illustrated in Figure~\ref{fig:singular}, the representations show a slight improvement when $t$ is small, but they quickly deteriorate as $t$ increases due to a rapid reduction in singular values and the effective rank number.
It indicates that the feature quality grows earlier in the reverse process, while gradually becoming uninformative in the later stages. 

\begin{figure}
    \centering
 \includegraphics[width=\linewidth]{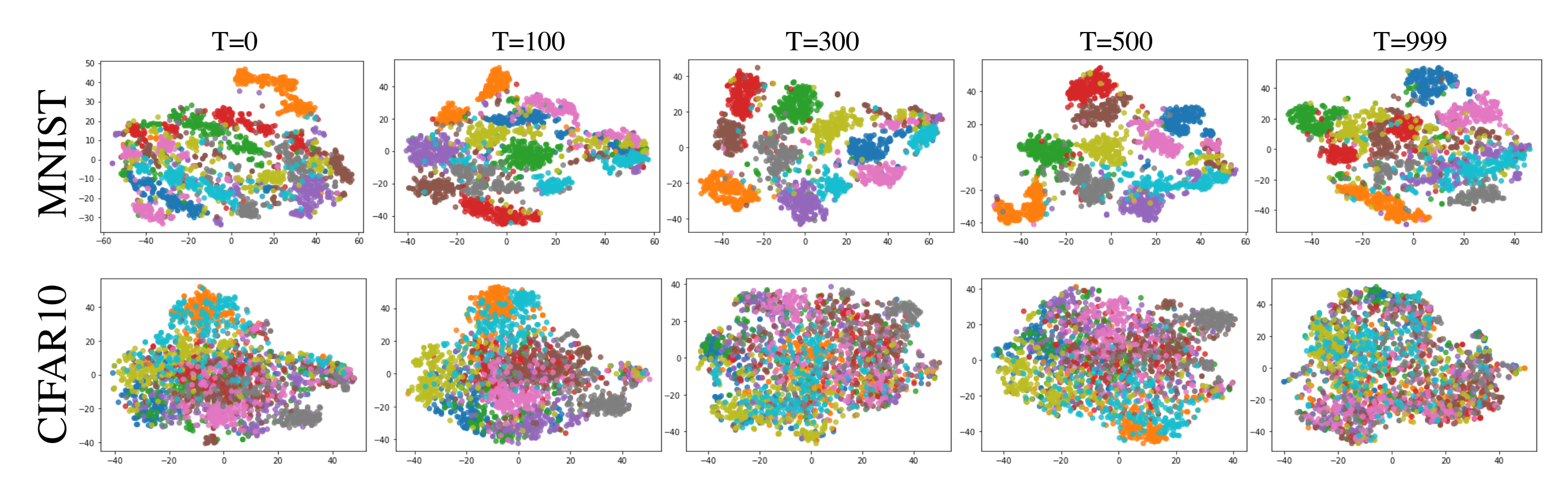}
 \vspace{0.5mm}
    \caption{T-SNE feature visualization on for unconditional DDPM trained on MNIST and CIFAR-10 datasets.}
    \label{fig:t-sne}
\end{figure}

 \noindent\textbf{Latent Embedding Visualization.} Visualizing the features using T-SNE for unconditional DDPMs, we observed a time-varying pattern. Figure~\ref{fig:t-sne} displays the data embeddings obtained through T-SNE for unconditional DDPMs. Our results reveal that, even without supervision, DPMs are able to group input samples at properly selected timestep, particularly $t=300$ on MNIST and $t=100$ on CIFAR-10. However, when $t$ is too small or large, the class-wise representations become inseparable, leading to less distinct clustering structures and blurred decision boundaries.

\begin{figure}

    \centering
    \includegraphics[width=\linewidth]{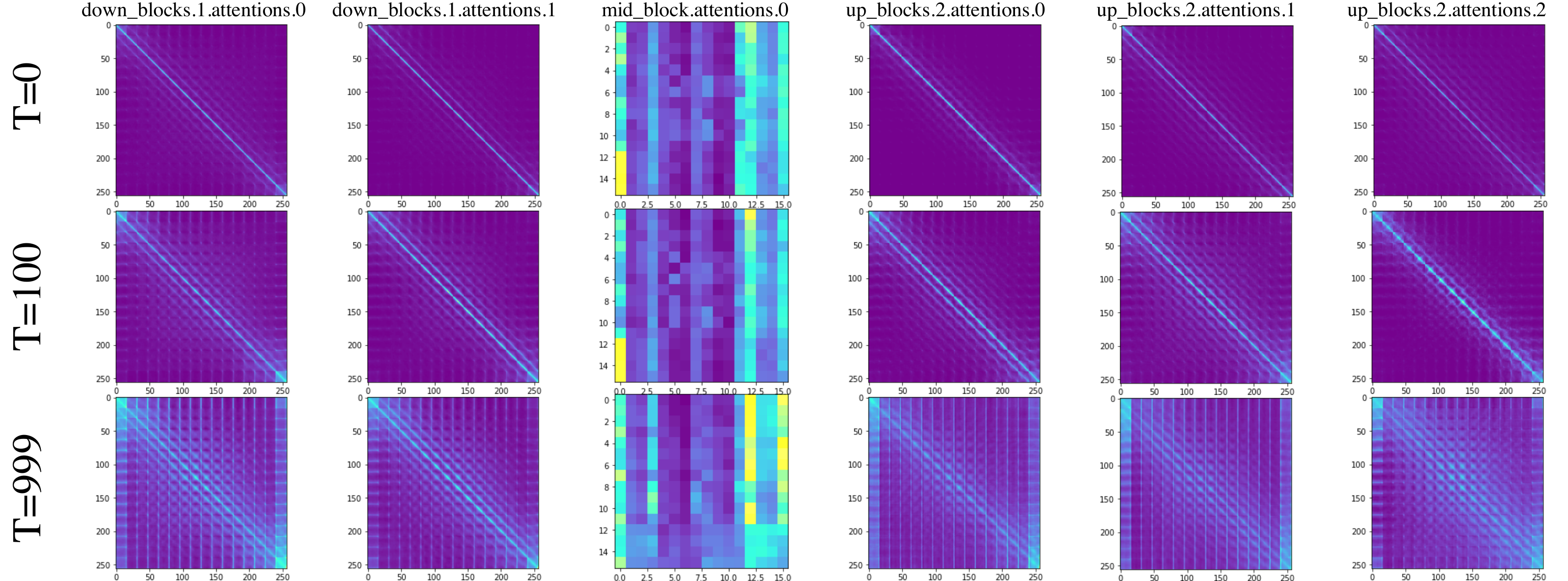}
    \vspace{0.5mm}
    \caption{Averaged self-attention map visualization on different feature layers with $T \in \{0, 100, 999\}$ on CIFAR-10.}
    \label{fig:attn}
\end{figure}

\noindent\textbf{Attention Patterns.} We also examined the attention patterns of DPMs on the CIFAR-10 dataset for different noise magnitudes corresponding to different scales of feature extraction. We calculate the self-attention map at $T\in \{0,100,999\}$ by averaging across 128 samples and all attention heads. The pixel value at the $i$-th column and $j$-th row refers to the attention weights between the $i$-th and the $j$-th token. As shown in  Figure~\ref{fig:attn}, higher noise levels led to the network attending to more image tokens, capturing long-range dependencies. Conversely, smaller $t$ corresponded to local attention, capturing fine-grained patterns.


To sum up, DPMs encompass a  latent space that evolves over time. Notably, the features at a moderately small timestep are found to possess the best geometric structure, potentially leading to optimal performance in the recognition task. Nevertheless, this optimal time step varies across various tasks and datasets, thereby posing significant challenges for applying trained DPM in non-generative tasks.
 
\textbf{}


\section{\name}
\label{sec:repfussion}

\begin{figure*}
    \centering
    \includegraphics[width=\linewidth]{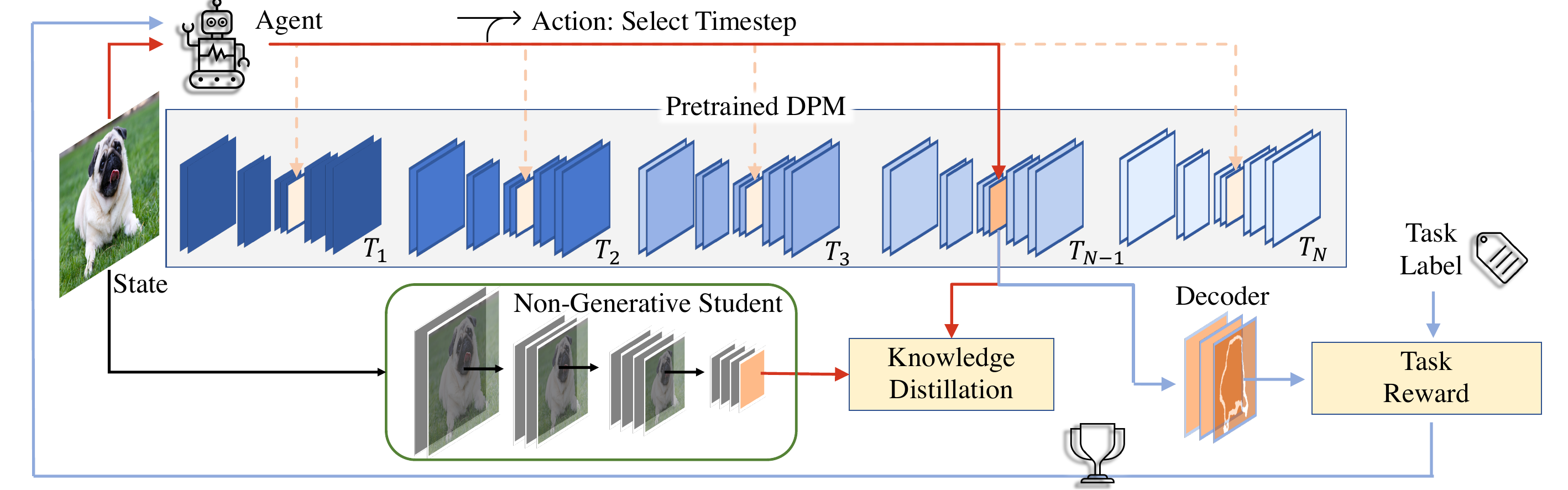}
    \caption{Overall pipeline for the~\name. Given an input sample,  we adopt REINFORRCE algorithm~\cite{williams1992simple} to determine the optimal time-step for knowledge distillation. }
    \label{fig:pipeline}
\end{figure*}

Given a pre-trained denoising network $\bm s(\cdot, \cdot; \bm\theta^*)$, our goal is to reuse its knowledge for recognition tasks.
 This can be achieved by distilling a recognition encoder $f(\cdot; \bm \theta_{f})$ from a DPM teacher. However, as discussed in Section ~\ref{sec:dpm_are_ae}, determining the ideal time step for knowledge transfer is non-trivial. To address this issue, we introduce a reinforcement learning approach to solve the selection problem. Our pipeline is illustrated in Figure~\ref{fig:pipeline}.




\subsection{Distill Knowledge from Diffusion Teacher}
We intend to distill the intermediate representation from a pre-trained diffusion model to a recognition student. Given an input $\mathbf{x}$ and its label $y$, we extract the feature pair from both the diffusion model at timestep $t$ and student model respectively, e.g. $\mathbf{z}^{(t)}\leftarrow \bm s(\mathbf{x}, t; \bm\theta^*)$  and $\mathbf{z} \leftarrow f(\mathbf{x}; \bm\theta_f)$.  We minimize the distance between $\mathbf{z}^{(t)}$ and $\mathbf{z}$ with respect to a loss function $\mathcal{L}_{\text{kd}}$
\begin{align}
    \min_{\bm\theta_f} \mathbb{E}[\mathcal{L}_{\text{kd}}(\mathbf{z}^{(t)},\mathbf{z})]\label{eq:distill}
\end{align}
In the experiment, we show that arbitrary distillation function $\mathcal{L}_{\text{kd}}$ can effectively boost the model's performance. Our implementation employs L2 distance~\cite{romero2014fitnets}, attention transfer~\cite{zagoruyko2016paying}, and relational knowledge distillation~\cite{park2019relational}.  

After the distillation phase, the student is reapplied as a feature extractor and fine-tuned on with the task label
\begin{align}
    \min_{\bm\theta_f} \mathbb{E}[\mathcal{L}_{\text{task}}(y,\hat{y})], \quad\text{where $\hat{y} = f(\mathbf{x};\bm\theta_f)$}
\end{align}
Here, $\hat{y}$ represents the student model's prediction and $\mathcal{L}_{\text{task}}$ is the task loss function. In this way, the distilled model using Eq.~\ref{eq:distill} is utilized to initialize the task training. As discussed, the selection of an optimal value for $t$ is a critical aspect of our pipeline. We provide a detailed explanation of our selection strategy for $t$ in Section~\ref{sec:reinforced}.

\subsection{Reinforced Time Selection for Distillation}\label{sec:reinforced}

Applying DPM for knowledge distillation presents a noteworthy challenge due to the mismatch between a group of time-indexed teachers and a single student network. This section outlines our criteria for time-step selection and explains how we optimize it through reinforcement learning.

Intuitively, we seek to extract the most informative feature, akin to a "golden standard", that can guide the student. To achieve this, we identify the optimal $t^*$ for each sample $\mathbf{x}$, such that the DPM's feature $\mathbf{z}^{(t^*)}$ is most predictive for $y$, thereby facilitating knowledge distillation. The realization of this objective leads to a nested optimization problem:
\begin{align}
    t^* &=\underset{t \in [0, T]}{\mathrm{argmin}}
 \Big\{ \inf_{\bm\theta_g} \mathcal{L}_{\text{task}}\big(y, g(\mathbf{z}^{(t)};\bm\theta_g)\big) \Big\}
 \label{eq:selection}
\end{align}
Here, $g(\cdot;\bm\theta_g)$ refers to an additional decoder that maps the feature of the DPM to the label space. While its generally impractical to train $g$ on all $(t,\mathbf{x})$ combinations\footnote{Gradient-based optimizations are not feasible due to the discrete nature of $t$.} exhaustively,  we have developed an efficient solution for determining the optimal distillation strategy using the REINFORCED algorithm~\cite{williams1992simple}.

Specifically, we define the state space as the input sample $\mathbf{x}$, and the action space as timestep $t$. We set the reward function $\mathcal{R}_{\mathbf{x}}^t$ to be the negative task loss $-\mathcal{L}_{\text{task}}\big(y, g(\mathbf{z}^{(t)};\bm\theta_g)\big)$. A policy network $\bm \pi_{\bm \theta_\pi}(t|\mathbf{x})$ is trained to determine, on a per-sample basis, which timestep should be evaluated. The network parameters $\bm\theta_\pi$ are optimized to maximize the reward function
\begin{align}
    \max_{\bm\theta_\pi} J(\bm\theta_\pi) = \mathbb{E} [\sum_{t} \bm\pi_{\bm\theta_\pi}(t|\mathbf{x}) \mathcal{R}_{\mathbf{x}}^t]+\lambda_H H(t)
    \label{eq:reinforced}
\end{align}
where $H(t)$~\cite{mnih2016asynchronous} is an entropy term to promote action diversity, and $\lambda_H=0.1$ is its weighting factor.

Besides, the decoder $g$ is trained jointly with the policy network to minimize task loss and improve the prediction accuracy of the ground-truth label. As such, we are optimizing Eq.~\ref{eq:selection} equivalently.

By assembling all the components, we present the~\name~algorithm in Figure~\ref{fig:pipeline}. In each training step, we begin by sampling the time from the action distribution $t\sim \text{Categorical}_T(\bm\pi_{\bm\theta_\pi}(t|\mathbf{x}))$ and subsequently extract the corresponding feature map $\mathbf{z}^{(t)}$ for each input sample. The student network is optimized to replicate the teacher's representation using Eq.~\ref{eq:distill}. Meanwhile, the policy network and teacher decoder are updated according to Eq.~\ref{eq:selection} and Eq.~\ref{eq:reinforced}. Despite its simplicity, we observe that this pipeline yields impressive results across various tasks.

\section{Experiment}

This section demonstrates that our \textbf{RepFusion} can be effectively utilized in a variety of recognition tasks with competitive performance, despite being learned from generative DPMs. The distilled encoders are evaluated on a range of downstream applications, including semantic segmentation, keypoint detection, and image classification. More experiments can be found in the supplementary materials.

\subsection{Experimental Setting}
\noindent\textbf{Datesets and Pre-trained DPMs.} We evaluate the proposed method on 3 tasks  across 4 different visual datasets. Those datasets include CelebAMask-HQ for semantic segmentation, WFLW for face landmark detection and CIFAR-10 and Tiny-ImageNet for image classification. The \textbf{CelebAMask-HQ}~\cite{CelebAMask-HQ} dataset is a collection of 30,000 high-resolution face images. 
Each image comes 
with its segmentation mask 
of 19 facial attributes. 
The \textbf{WFLW} dataset~\cite{wayne2018lab} consists of 10,000 faces that have been annotated with 98 landmarks, 
with 7,500 for training and 2,500 for testing. \textbf{CIFAR-10}~\cite{Krizhevsky09learningmultiple} is an image dataset containing $32\times 32$ images from 10 classes, 
and is split into 50,000 images for training and 10,000 for validation. \textbf{Tiny-ImageNet}~\cite{le2015tiny}  comprises images
from 200 classes 
downsized to $64\times 64$. Specifically, each class contains 500 training, 50 validation, and 50 test images.

We have gathered open-source DPMs models on diffusers\footnote{https://huggingface.co/docs/diffusers/index} and guided diffusion\footnote{https://github.com/openai/guided-diffusion}. To perform knowledge distillation, we utilize ADM~\cite{dhariwal2021diffusion} model that was trained on $64\times 64$ ImageNet, as the teacher model for TinyImageNet. For face parsing, we use DDPM~\cite{ho2020denoising}  trained on CelebA-HQ, as the teacher model. We do not apply distillation on WFLW, but direct transfer the weight learned from the CelebA-HQ segmentation as initialization. Furthermore, we reuse the DDPM that was trained on unconditional CIFAR-10 as the teacher model for the same dataset.


\noindent\textbf{Distillation.} We utilize the pipeline described in Section~\ref{sec:repfussion} to train the networks ${f,g,\pi}$ for 200 epochs with a mini-batch size of 128 across all tasks. The networks are optimized using SGD with an initial learning rate of 0.1, a momentum term of 0.9, and a weight decay of 1e-4. We apply learning rate decay at the 100th and 150th epochs. During training, the only augmentation applied to training images is random horizontal flipping, and no pre-trained weights are loaded by default.
Feature distillation is performed using the embeddings extracted from the \textsc{Mid-Block} of the DPM and the last layer of the student network. A hint loss is used for feature distillation by default, and this selection will be verified in the ablation study. Furthermore, we define $g$ as a linear layer for classification and FCN~\cite{long2015fully} for segmentation. The policy network $\bm\pi$ is a 3-layer convolutional net followed by 3 linear layers.

\noindent\textbf{Segmentation.} We use BiSeNetv1~\cite{yu2018bisenet} with ResNet-18 and ResNet-50 backbones as the baseline for semantic segmentation. The networks are distilled on CelebA-HQ at $256\times 256$ resolution, then trained for 160k iterations on segmentation labels using SGD optimizer with batch size 16, initial learning rate 0.01, momentum 0.9, and polynomial learning rate scheduler. Random cropping of size $448\times 448$ is applied during training. Finally, we evaluate the performance of the model on the test set at a scale of $512\times 512$ and report its performance based on the mean Intersection over Union (mIOU), mean accuracy of each class (mAcc), and all-pixel accuracy (aAcc).

\noindent\textbf{Landmark Detection.} The distilled segmentation student on CelebA-HQ is reused for Top-down heatmap-based keypoint estimation~\cite{xiao2018simple} with HRNetv2-w18~\cite{wang2020deep} and ResNet-50. The model was trained for 60 epochs with a batch size of 128 using the Adam optimizer with an initial learning rate of 2e-3. Images were resized to 256 during both training and testing. Performance was evaluated on using the normalized mean error (NME) metric on 7 subsets.

\begin{table}[]
    \centering
    \renewcommand{\arraystretch}{1.2}
    \setlength{\tabcolsep}{5pt}
        \caption{Semantic segmentation results on CelebAMask-HQ~\cite{CelebAMask-HQ}, compared with hint-based knowledge distillation~\cite{romero2014fitnets}.}
    \label{tab:distill_seg}
    \begin{tabular}{l|c|c|c|c}
    \hline
        Teacher & Student & mIOU & aAcc & mAcc \\
         \hline
         - & ResNet-18  &61.22 & 94.65 & 70.29  \\
         ResNet-18 & ResNet-18 & 62.73 & 94.46 & 71.88  \\
          ResNet-50 & ResNet-18 & 63.12 & 94.72 &72.44 \\
         \name(ours) & ResNet-18 & \textbf{67.37} & \textbf{94.75} & \textbf{76.37} \\
        \hline
        - & ResNet-50 & 65.81 & 94.83 & 75.31\\
        ResNet-50 & ResNet-50 & 68.21 & 94.96 &77.23 \\
         \name(ours) & ResNet-50 & \textbf{70.63} & \textbf{95.24} & \textbf{80.08}\\
         \hline
    \end{tabular}
\end{table}
\begin{table}[]
\renewcommand{\arraystretch}{1.2}
    \centering
        \caption{Semantic segmentation results on CelebAMask-HQ~\cite{CelebAMask-HQ}, compared with self-supervised learning approaches.}
    \label{tab:selsup_seg}
    \setlength{\tabcolsep}{9pt}
    \begin{tabular}{l|c|c|c}
    \hline
        Method &  mIOU & aAcc & mAcc \\
         \hline
          ResNet-18 &61.22 & 94.65 & 70.29  \\

        + MoCov2~\cite{chen2020improved} & 66.38 & \textbf{94.84} & 74.23 \\
        + SwAV~\cite{caron2020unsupervised} & 67.14 & 94.66 & 75.92 \\
        + DeepClusterv2~\cite{caron2020unsupervised} & 66.92 & 94.51 & 76.11 \\ \hline
        + \name(ours) & \textbf{67.37} & 94.70 & \textbf{76.37} \\
        \hline\hline
     ResNet-50 &  65.81 & 94.83 & 75.31\\
        + MoCov2~\cite{chen2020improved} & 69.81 & 95.32 & 79.55 \\
        + SwAV~\cite{caron2020unsupervised} & 70.41 &  95.26 & 79.40 \\
        + DeepClusterv2~\cite{caron2020unsupervised} & 68.83 & 95.27 & 79.43 \\
        \hline
        + \name(ours) & \textbf{70.89} & \textbf{95.47} & \textbf{80.14} \\
         \hline
    \end{tabular}
\end{table}
\noindent\textbf{Classification.} On CIFAR-10 and Tiny-ImageNet, we follow the conventional setting and apply ResNet-18~\cite{he2016deep}, WideResNet~(WRN) 28-2~\cite{zagoruyko2016wide} and MobileNetv2~(MBNv2)~\cite{sandler2018mobilenetv2} as the student models. 

\noindent\textbf{Baselines.} We compared our method to several self-supervised learning approaches, including MoCov2~\cite{chen2020improved}, SwAV~\cite{caron2020unsupervised}, and DeepCluster-v2~\cite{caron2020unsupervised}. To ensure a fair comparison, we used the official implementation and pre-trained the models directly on CelebA-HQ without large-scale pretraining, keeping all hyper-parameters the same as in their respective papers.
However, we did not compare with DatasetGAN~\cite{zhang2021datasetgan} and DatasetDDPM~\cite{baranchuk2021label} as they are primarily designed for few-shot segmentation. 

Besides, we assessed our performance  by comparing it with the other distillation techniques. Specifically, we initially trained a teacher model from scratch and distilled a student model by mimicking the teacher's features. 

\subsection{Evaluation on Recognition Tasks}

\subsubsection{Semantic Segmentation}
To evaluate the effectiveness of our method, we conducted a comparative analysis on CelebAMask-HQ by comparing it with two distinct lines of research, namely (1) knowledge distillation from a supervised teacher and (2) self-supervised representation learning models.

\noindent\textbf{Quantitative Results.} We show the numerical results for face parsing with distilled and self-supervised models in Table~\ref{tab:distill_seg} and Table~\ref{tab:selsup_seg}. First, our approach significantly outperforms the knowledge distillation counterparts. Precisely, while raw ResNet-18 achieved an mIOU score of 61.22, our method achieved an mIOU of 67.37. This represents an improvement of 6.15 over the baseline model and a 4.25 improvement over the distilled student from ResNet-50. Our \name~also matches the performance with up-to-date self-supervised representation learner. 
These findings underscore the notable improvement in the segmentation task through knowledge distillation from off-the-shelf DPMs.

\begin{figure}
    \centering
    \includegraphics[width=\linewidth]{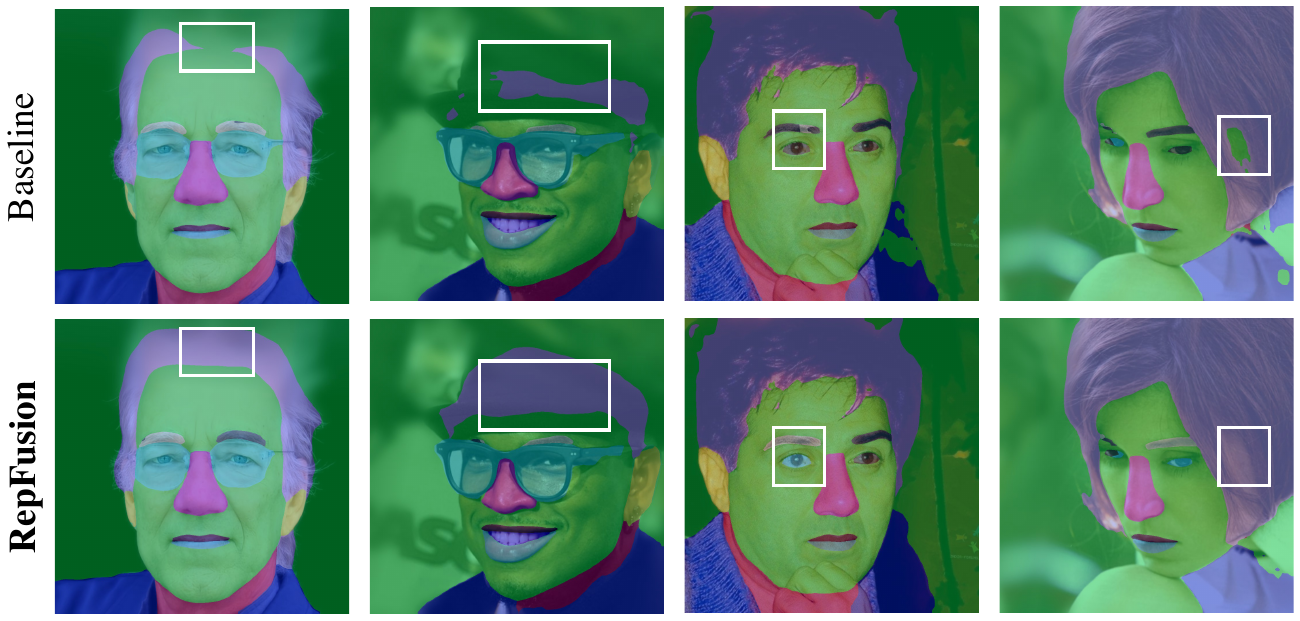}
    \vspace{1mm}
    \caption{\textbf{Face parsing} results on CelebAMask-HQ dataset. \name~excels on segmenting rare patterns and minority categories. }
    \label{fig:seg_mask_vis}
    \vspace{1mm}
\end{figure}

\begin{figure}
    \centering
    \includegraphics[width=\linewidth]{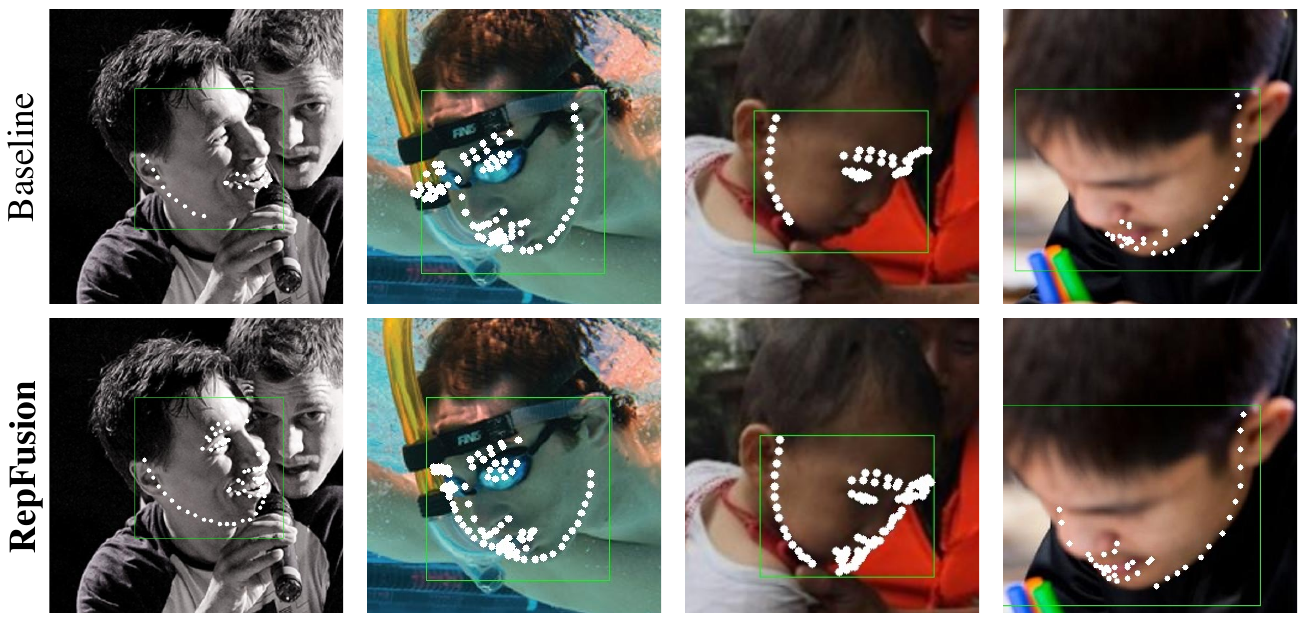}
    \vspace{1mm}
    \caption{\textbf{Face keypoint detection} results on WFLW~\cite{wayne2018lab} dataset. \name~can successfully handle the hard cases (large pose and low image quality).}
    \label{fig:face_det}
     \vspace{1mm}
\end{figure}

\noindent\textbf{Qualitative Results.} Additionally, we visualize the segmented facial attribute with ResNet-50 model in Figure~\ref{fig:seg_mask_vis}, where we highlight the errors made by the baseline model using white rectangles. It is evident that the baseline model struggles to accurately segment rare patterns and minority categories such as \texttt{white hair}, \texttt{hat} and distinguish between the \texttt{left eyebrows} and \texttt{right eyebrows}.  By distilling from DPMs, our~\name~enables the model to capture fine-grained details to better discern between these features, resulting in more accurate segmentation.

\begin{table*}[ht]
\renewcommand{\arraystretch}{0.9}
\caption{Normalized Mean Error~(NME) on WFLW for head keypoint detection. A smaller value indicates better performance.}
\label{tab:keypoint_face}
\renewcommand{\arraystretch}{1.2}
\setlength{\tabcolsep}{2.5pt}
    \centering
    \begin{tabular}{l|l|l|l|l|l|l|l}
    \hline
        Method & $\textit{NME}_{\text{test}}\downarrow$ & $\textit{NME}_{\text{pose}}\downarrow$ & $\textit{NME}_{\text{illumination}}\downarrow$ & $\textit{NME}_{\text{occlusion}}\downarrow$ & $\textit{NME}_{\text{blur}}\downarrow$ & $\textit{NME}_{\text{makeup}}\downarrow$  & $\textit{NME}_{\text{expression}}\downarrow$ \\ \hline
        HRNetv2w18 & 4.33 & 7.81 &	4.42&	5.35&	4.92&	4.36 &	4.55\\ 
        + MoCov2~\cite{chen2020improved} & 4.27 &	7.64	& 4.25 &	5.22 &	4.86&	4.22 & 4.49	\\ 
        + SwAV~\cite{caron2020unsupervised} & 4.26 &	7.54	& 4.19 &	5.23&	4.82&	4.19 & 4.49	\\ 
        + DeepClusterv2~\cite{caron2020unsupervised} & 4.31 &	7.58	& 4.22 &	5.19&	4.84&	4.23 & 4.48	\\ \hline
        + \name & \textbf{4.23} &	\textbf{7.42} &	\textbf{4.18}&	\textbf{5.10}&	\textbf{4.80} &	\textbf{4.15}& 	\textbf{4.48} \\ \hline\hline
        ResNet-50 & 4.58&	8.15&	4.58&	5.63&	5.23&	4.54&	5.01 \\ 
        + MoCov2~\cite{chen2020improved} & 4.49 &	7.99	& 4.52 &	5.49 &	5.14 &	4.42 & 4.85	\\ 
        + SwAV~\cite{caron2020unsupervised} & 4.47 &	7.97	& 4.53 &5.49&	5.12 &	4.40 & 4.87	\\ 
        + DeepClusterv2~\cite{caron2020unsupervised} & 4.51 &	8.00	& 4.55 &	5.51 &	5.18 &	4.39 & 4.88	\\ 
         \hline
        + \name & \textbf{4.47} &	\textbf{7.97} &	\textbf{4.49} &	\textbf{5.48} &	\textbf{5.10} &	\textbf{4.39} &\textbf{4.84} \\ \hline
    \end{tabular}
\end{table*}



\subsubsection{Keypoint Detection}
In this study, we compare our distill image encoder on CelebA-HQ to other self-supervised learning approaches for facial keypoint detection.\\
\noindent\textbf{Quantative Results.} We present the results in Table~\ref{tab:keypoint_face}. All of the self-supervised methods yielded performance improvements. However, our proposed approach outperformed the other methods, achieving the smallest mean errors for both ResNet-50 and HRNetv2w18.

It is noteworthy that the distillation process is proved to be particularly beneficial in challenging scenarios characterized by large pose variations, heavy occlusions, and extreme illumination. In such scenarios, our proposed method achieved significant improvements of 0.39/0.18 on $\textit{NME}_{\text{pose}}$, 0.25/0.15 on $\textit{NME}_{\text{occlusion}}$, and 0.24/0.09 on $\textit{NME}_{\text{illumination}}$ scores. These findings highlight the effectiveness of our proposed distillation approach in enhancing the model's performance in challenging scenarios.

\noindent\textbf{Qualitative Results. } Figure~\ref{fig:face_det} showcases the landmark detection visualization obtained by ResNet-50. The outcomes expose that the baseline model, lacking distillation, encounters challenges in recognizing faces with extensive head movements or off-angle orientations. In contrast, our proposed technique successfully identifies landmarks that are otherwise unobservable. These findings validate the efficacy of our approach in augmenting the model's recall in detecting facial landmarks, especially in complex scenarios characterized by notable pose variations, occlusions, and suboptimal image quality.

\subsubsection{Image Classification}
In this section, we aim to compare the effectiveness of learned representations obtained through distillation from DPMs by assessing the classification performance~\cite{hjelm2018learning,he2020momentum} on widely used benchmark datasets, namely CIFAR-10 and TinyImageNet. We compare our approach with typical distillation schemes on top of Hint-based~\cite{romero2014fitnets}, Attention Transfer-(AT)~\cite{zagoruyko2016paying}, and Relational Knowledge Distillation~(RKD)~\cite{park2019relational} losses. 

\begin{table}[!tb]
\renewcommand{\arraystretch}{1.2}
\setlength{\tabcolsep}{1pt}
    \centering
    \vspace{2mm}
     \footnotesize
    \caption{Test Accuracy (\%) comparison on CIFAR-10.}
    \label{tab:CIFAR-10}
    \begin{tabular}{c|c|c|c|c||>{\columncolor[gray]{0.9}}c}
    \hline
    Teacher:Acc & Student:Acc & Hint~\cite{romero2014fitnets} & AT~\cite{zagoruyko2016paying} & RKD~\cite{park2019relational}  & Avg Imp.  \\
   \hline
    ResNet-18:94.47 & \multirow{4}{*}{ResNet-18:94.47} & 95.06 & 94.66 &94.90 & \textcolor{gray}{+0.40}  \\
    WRN28-2:94.15 &  & 94.87 & 94.63 &94.57  & \textcolor{gray}{+0.22}\\
    WRN28-10:95.18 &  & 94.95 & 94.35 &95.04 & \textcolor{gray}{+0.31} \\
    \name(Ours) & & 95.01 &95.09 &94.58 & \textcolor{Green}{\textbf{+0.42}}\\ 
     \hline
      ResNet-18:94.47& \multirow{4}{*}{WRN28-2:94.15} & 94.39 & 94.42 & 94.57 & \textcolor{gray}{+0.31} \\
      WRN28-2:94.15&  & 94.53 &94.12 & 94.07 & \textcolor{gray}{+0.09} \\
       WRN28-10:95.18 &  & 94.35 & 94.05 & 94.61 & \textcolor{gray}{+0.18} \\
       \name(Ours) & & 94.52 & 94.42 & 94.66 &  \textcolor{Green}{\textbf{+0.38}}\\
     \hline
    \end{tabular}
    \vspace{2mm}
\end{table}

\noindent\textbf{Results.} Table~\ref{tab:CIFAR-10} and Table~\ref{tab:tinyimagenet} present a comparison of the test accuracy of various models on the CIFAR-10 and Tiny-ImageNet datasets. On CIFAR-10, our proposed method performs comparably to other teachers in terms of the average increase in test accuracy, exhibiting a gain of +0.42/+0.38 compared to its student. On Tiny-ImageNet, our proposed technique achieves the highest average accuracy gain, with an improvement of +1.83 over ResNet-18 and +0.42 on MBNv2. These results highlight the effectiveness of our approach in improving the performance of the student models on recognition tasks. 

Furthermore, all distillation losses (Hint, AT, and RKD) lead to a boost in student model accuracy, with hint-based distillation showing the best improvement. Notably, our model outperforms student models that were distilled from supervised teachers, establishing its efficacy in acquiring valuable representations for classification tasks.

\textbf{Comparison Among Tasks.} In our experiments, we found that when distilling from the DPM teacher, there was a notable performance improvement in segmentation compared to a more modest enhancement in classification. This suggests that the features learned by DPMs are more effective for tasks that prioritize \emph{local information} over \emph{global semantics}. This conclusion is supported by findings from other recent studies~\cite{li2023dreamteacher, zhao2023unleashing}.
\begin{table}[!tb]
\renewcommand{\arraystretch}{1.2}
\setlength{\tabcolsep}{1pt}
    \centering
    \vspace{2mm}
     \footnotesize
    \caption{Test Top-1 Accuracy (\%) comparison on Tiny-ImageNet.}
    \label{tab:tinyimagenet}
    \begin{tabular}{c|c|c|c|c||>{\columncolor[gray]{0.97}}c}
    \hline
    Teacher:Acc & Student:Acc & Hint~\cite{romero2014fitnets} & AT~\cite{zagoruyko2016paying} & RK~\cite{park2019relational}  & Avg Imp.   \\
   \hline
    ResNet-18:60.37 & \multirow{4}{*}{ResNet-18:60.37} & 61.48 & 61.93 &61.06 &\textcolor{gray}{+1.12}  \\
    WRN28-2:59.30 &  & 61.91 & 61.11 &61.31 & \textcolor{gray}{+1.07} \\
    WRN28-10:65.73 &  & 61.65 & 61.12 &62.08  &\textcolor{gray}{+1.25} \\
    \name(Ours) & & 62.16 &62.05 & 62.38 & \textbf{\textcolor{Green}{+1.83}}\\ 
     \hline
    ResNet-18:60.37& \multirow{4}{*}{MBNv2:61.69}  & 61.99 & 61.84 & 62.40 & \textbf{\textcolor{gray}{+0.42}}\\
      WRN28-2:59.30&  & 61.97 & 61.20 & 62.37  & \textcolor{gray}{+0.20}\\
       WRN28-10:65.73 &  & 61.81 & 61.63 &61.63 & \textcolor{gray}{+0.04}\\
       \name(Ours) & & 61.93 &62.01 & 62.27 &\textbf{\textcolor{Green}{+0.42}} \\
     \hline
    \end{tabular}
    \vspace{2mm}
\end{table}

     
\subsection{Ablation Study and Analysis on Time Selection}
This section validates the necessity and effectiveness of our time-step selection approach. We specifically investigate its importance and the achieved outcomes.

\noindent\textbf{Do We Need Time-step Selection?} To assess the necessity of the reinforced strategy in selecting time steps, we conducted experiments using alternative selection methods, namely \emph{random} and \emph{manual} selection. For the \emph{random} strategy, a time-step is sampled uniformly from the set of time-steps $\{0,\dots, T\}$ for each data point. In contrast, the \emph{manual} selection method involves selecting a fixed time-step $t$ for all data points, with $t\sim\{0,10,50,100,200,300,500,999\}$. We validate the results on CIFAR-10 with the ResNet18 model.

\begin{figure}
\vspace{-2mm}
    \centering
\includegraphics[width=\linewidth]{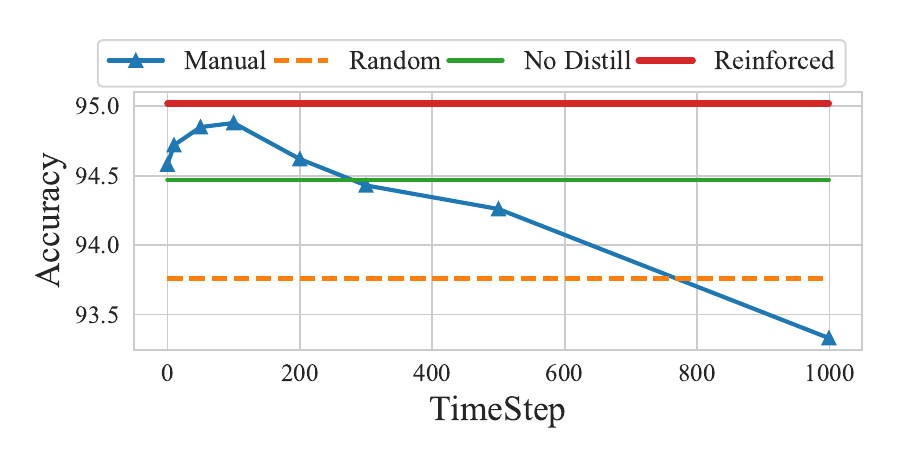}\vspace{-3mm}
    \caption{Ablative study on the time-step selection.}
    \label{fig:time_ablation}
\end{figure}

Figure~\ref{fig:time_ablation} presents the performance with different time selection strategies. The results indicate that distillation from randomly sampled or large fixed timesteps leads to a significant degradation in performance, even compared to the no-distill baseline. This finding aligns with the derivation in Section~\ref{sec:dpm_are_ae} that large timestep features are unlikely to provide discriminative information, leading to a negative impact on performance. While a fixed small $t$ approach can slightly improve performance, it is not universally effective. In contrast, our reinforced strategy that identifies the optimal $t$ on a per-sample basis outperforms the fixed $t$ approach.

\noindent\textbf{What Time Has Been Selected?} One natural question is what timestep is selected across various tasks. Our core observation reveals that different tasks give slightly varied outcomes. Initially, during training, the time step $t$ is randomly selected, but gradually it tends to converge to a narrow range of $0-200$. We have included the full results in the supplementary, owing to page limitations.

\section{Conclusion}
In this work, we demonstrate the efficacy of Diffusion Probabilistic Models (DPMs) in enhancing visual representations for recognition tasks. By establishing a connection between DPMs and denoising auto-encoders, we empirically validate the statistical properties of DPM-extracted features. However, utilizing these features for non-generative tasks poses challenges. To address this, we propose a novel knowledge distillation approach called \name, which leverages a reinforcement learning framework to determine the optimal timing for representation distillation. Our method demonstrates consistent improvements across various recognition benchmarks, highlighting the remarkable potential of DPMs in learning effective representations and facilitating transfer to downstream tasks.
\vspace{-2mm}
\section*{Acknowledgment}
\vspace{-2mm}
This research is supported by the Ministry of Education, Singapore, under its Academic Research Fund Tier 2
(Award Number: MOE-T2EP20122-0006).

{\small
\bibliographystyle{ieee_fullname}
\bibliography{egbib}
}

\end{document}